\documentclass[10pt,table]{article} 
\usepackage[preprint]{tmlr}

\usepackage{amsmath,amsfonts,bm}

\def\eqref#1{equation~\ref{#1}}

\def\1{\bm{1}}

\DeclareMathAlphabet{\mathsfit}{\encodingdefault}{\sfdefault}{m}{sl}
\SetMathAlphabet{\mathsfit}{bold}{\encodingdefault}{\sfdefault}{bx}{n}

\usepackage{hyperref}
\usepackage{url}

\usepackage{graphicx}
\usepackage[normalem]{ulem}
\usepackage{booktabs}       
\usepackage{amsmath}
\usepackage{amsthm}

\newtheorem{theorem}{Theorem}[section]
\newtheorem{prop}[theorem]{Proposition}
\newcommand{\code}[1]{\texttt{#1}}
\definecolor{light-gray}{gray}{0.95}
\definecolor{pearDark}{HTML}{2980B9}

\title{Bag of Image Patch Embedding \\ Behind the Success of Self-Supervised Learning}

\author{\name Yubei Chen $^{1,6, *}$ \email yubeichen@nyu.edu
      \vspace{-0.1in} \AND 
      \name Adrien Bardes $^{1,2, *}$
      \vspace{-0.1in} \AND
      \name Zengyi Li $^{3,4}$
      \vspace{-0.1in} \AND Yann LeCun $^{1,5,6}$ \\ \\
      \addr $^1$\normalfont Meta AI - FAIR $^2$ Inria, \'{E}cole normale sup\'{e}rieure, CNRS, PSL Research University \\
      $^3$\normalfont Redwood Center $^4$Physics Dept., UC Berkeley\\
      $^5$\normalfont Courant Institute $^6$ Center for Data Science, New York University\\
      $^{*}$\normalfont Equal contribution
      }

\def\month{MM}  
\def\year{YYYY} 
\def\openreview{\url{https://openreview.net/forum?id=XXXX}} 

\begin{document}

\maketitle

\begin{abstract}
Self-supervised learning (SSL) has recently achieved tremendous empirical advancements in learning image representation. However, our understanding of the principle behind learning such a representation is still limited. This work shows that joint-embedding SSL approaches primarily learn a representation of image patches, which reflects their co-occurrence. Such a connection to co-occurrence modeling can be established formally, and it supplements the prevailing invariance perspective. We empirically show that learning a representation for fixed-scale patches and aggregating local patch representations as the image representation achieves similar or even better results than the baseline methods. We denote this process as {\it BagSSL}. Even with $32\times 32$ patch representation, BagSSL achieves $62\%$ top-1 linear probing accuracy on ImageNet. On the other hand, with a multi-scale pretrained model, we show that the whole image embedding is approximately the average of local patch embeddings. While the SSL representation is relatively invariant at the global scale, we show that locality is preserved when we zoom into local patch-level representation. Further, we show that patch representation aggregation can improve various SOTA baseline methods by a large margin. The patch representation is considerably easier to understand, and this work makes a step to demystify self-supervised representation learning.
\end{abstract}

\section{Introduction}
In many application domains, self-supervised representation learning experienced tremendous advancements in the past few years. In terms of the quality of learned features, unsupervised learning has caught up with supervised learning or even surpassed the latter in many cases. This trend promises unparalleled scalability for data-driven machine learning in the future. One of the most successful self-supervised learning paradigms is joint-embedding SSL \citep{wu2018unsupervised, chen2020simple, grill2020BYOL, chen2020improved, bardes2021vicreg, zbontar2021barlow, yeh2021decoupled, li2022neural}, which uses a Siamese network architecture \citep{bromley1993signature}. These methods follow a general goal: to transform different views (augmentation) of the same instance (image) closer in the representation space, and meanwhile, the new space is not collapsed, or in other words, important geometric and stochastic structures are preserved. 

While we celebrate the empirical success of SSL, our understanding of this learning process still needs to be improved. This work shows that {\bf \em{joint-embedding SSL approaches are primarily learning a representation of image patches, which reflects their co-occurrence.}} To demonstrate this, we first establish a formal connection between joint-embedding SSL and co-occurrence modeling. Then, we show that learning a representation for fixed-scale patches and linearly aggregating patch representations (bag-of-local-features) as the image representation achieves similar or even better results than learning the representation with multi-scale crops. Empirical results are shown with several SSL methods\citep{chen2020simple, bardes2021vicreg, li2022neural, grill2020BYOL} on CIFAR10, CIFAR100, ImageNet100, and ImageNet-1K datasets. Even with $32\times 32$ patch representation, we can achieve $62\%$ top-1 linear probing accuracy on ImageNet-1K. And KNN classifier also works surprisingly well with the aggregated patch feature. These findings resonate with recent works in supervised learning based on local features \citep{brendel2018BagNet, dosovitskiy2020image, trockman2022patches}. 
We further show that for baseline SSL methods pretrained with multi-scale crops, the whole-image representation is approximately a linear aggregation of local patch embeddings. The local patch representation is considerably easier to understand, and we provide visualization to show that the representation space preserves locality when we zoom into the local patch representation. These discoveries supplement the prevailing invariance perspective, provide useful understanding for the success of joint-embedding SSL, and make a step to demystify self-supervised representation learning.

\section{Related Works}

{\bf Joint-embedding SSL: Invariance without Collapse.} 
Joint-embedding SSL (or instance-based SSL) \citep{wu2018unsupervised} views each of the images as a different class and uses data augmentation \citep{DosovitskiyFSRB16} to generate different views from the same image. As the number of classes equals the number of images, it is formulated as a massive classification problem, which may require a huge buffer or memory bank. Later, SimCLR \citep{chen2020simple} simplifies the technique significantly and uses an InfoNCE-based formulation to restrict the classification within an individual batch. At the same time, it's widely perceived that contrastive learning needs the ``bag of tricks,'' e.g., large batches, hyperparameter tuning, momentum encoding, memory queues, etc. Later works \citep{chen2021simsiam,yeh2021decoupled, haochen2021provable} show that many of these issues can be easily fixed. Recently, several even simpler non-contrastive learning methods\citep{bardes2021vicreg, zbontar2021barlow,li2022neural} are proposed, where one directly pushes the representation of different views from the same instance closer while maintaining a non-collapsing representation space. Joint-embedding SSL methods mostly differ in their means of achieving a non-collapsing solution. This include classification versus negative samples\citep{chen2020simple}, Siamese networks \citep{he2020MoCo,grill2020BYOL} and more recently, covariance regularization \citep{ermolov2021whitening, zbontar2021barlow, bardes2021vicreg, haochen2021provable,li2022neural,bardes2022vicregl}.
The covariance regularization has also long been used in many classical unsupervised learning methods \citep{roweis2000nonlinear, tenenbaum2000global, wiskott2002slow, chen2018sparse}, also to enforce a non-collapsing solution. There is a duality between the spectral contrastive loss\citep{haochen2021provable} and the non-contrastive loss, which we briefly discuss the intuition in Appendix \ref{app:duality_short}.

All previously mentioned joint-embedding SSL methods pull together representations of different views of the same instance. Intuitively, the representation would eventually be invariant to the transformation that generates those views. We would like to provide further insight into this learning process: The learning objective can be understood as using the inner product to capture the co-occurrence statistics of local image patches. We also provide visualization to study whether the learned representation truly has this invariance property.

{\bf Patch-Based Representation.}
Many works have explored the effectiveness of path-based image features. In the supervised setting, Bagnet\citep{brendel2018BagNet} and \citet{thiry2021patchinconvkernel} showed that aggregation of patch-based features could achieve most of the performance of supervised learning on image datasets. In the unsupervised setting, \citet{gidaris2020SSLbypredBOW} performs SSL by requiring a bag-of-patches representation to be invariant between different views. Due to architectural constraints, vision transformer-based methods also naturally use a patch-based representation \citep{he2021MAESSL,bao2021ImageBert}. 

{\bf Learning Representation by Modeling the Co-Occurrence Statistics.}
The use of word vector representation has a long history in NLP, which dates back to the 80s \citep{rumelhart1986learning, dumais2004latent}. Perhaps one of the most famous word embedding results, the word vector arithmetic operation, was introduced in \citet{mikolov2013efficient}. Particularly, to learn this embedding, a task called ``skip-gram'' was used, where one uses the latent embedding of a word to predict the latent embedding of the word vectors in a context. A refinement was proposed in \citet{mikolov2013distributed}, where a simplified variant of Noise Contrastive Estimation (NCE) was introduced for training the ``Skip-gram'' model. The task and loss are deeply connected to the SimCLR and its InfoNCE loss. Later, a matrix factorization formulation was proposed in \citet{pennington2014glove}, which uses a carefully reprocessed concurrence matrix compared to latent semantic analysis. While the task in Word2Vec and SimCLR is apparently similar, the underlying interpretations are quite different. In joint-embedding SSL methods, one pervasive perception is that the encoding network is trying to build invariance, i.e., different views of the same instance shall be mapped to the same latent embedding. This work supplements this classical opinion and shows that similar to Word2Vect, joint-embedding SSL methods can be understood as building a distributed representation of image patches by modeling the co-occurrence statistics.

\section{Joint-Embedding SSL Models Image Patch Co-Occurrence}
As we discussed earlier, recent works show that there exists a duality \cite{Garrido2023duality} between contrastive SSL and non-contrastive SSL methods, and these methods are essentially solving the same problem. So we only need to establish the connection between co-occurrence statistics modeling and one joint-embedding SSL method. In the following, we show that the spectral contrastive learning \cite{haochen2021provable} loss function models the co-occurrence statistics. We also replace multi-scale crop augmentation with fixed-scale image patches during SSL training. Thus the learning process is precisely modeling the patch-level co-occurrence statistics. After training, SSL has learned a representation for fixed-scale image patches. The whole-image representation is a linear aggregation of local patch representations.

{\bf Joint-embedding SSL and Co-Occurrence.} Let's assume $\vec{x}_1$ and $\vec{x}_2$ are two color-augmented patches sampled from the dataset. We denote their marginal distribution by $p(\vec{x}_1)$ and $p(\vec{x}_2)$, which includes variation due to sampling different images, spatial locations within an image, and random color augmentation. We denote their joint distribution by $p(\vec{x}_1,\vec{x}_2)$, which is the probability that they co-occur within the same image. $\vec{z}_1$ and $\vec{z}_2$ are corresponding embedding vectors of $\vec{x}_1$ and $\vec{x}_2$ in the representation space. The representation transform is parameterized by a neural network. The next proposition shows that the spectral contrastive loss function is equivalent to co-occurrence statistics modeling.

\begin{prop}
\label{prop:cooccurrence}
The spectral contrastive loss function $L_S$:
\begin{align}
L_{S} =  \mathbb{E}_{p(\vec{x}_1, \vec{x}_2)}\left[- \vec{z}_1^T\vec{z}_2\right] + \lambda\mathbb{E}_{p(\vec{x}_1)p(\vec{x}_2)} \left(\vec{z}_1^T\vec{z}_2\right)^2
\label{opt:spectral_reg}
\end{align}
is equivalent to the co-occurrence statistics modeling loss function $L_C$:
\begin{align}
L_{C} =  \int p(\vec{x}_1)p(\vec{x}_2) \left[ w\vec{z}_1^T\vec{z}_2 - \frac{p(\vec{x}_1, \vec{x}_2)}{p(\vec{x}_1)p(\vec{x}_2)} \right]^2 d\vec{x}_1 d\vec{x}_2
\label{opt:cooccurrence}
\end{align}
where $\lambda=\frac{w}{2}$, and $w$ is a constant weighting factor.
\end{prop}

The proof is rather straightforward and is presented in Appendix~\ref{app:proof_1}. The first term in $L_S$ ensures that co-occurring image patches are embedded closer in the embedding space and the second term in $L_S$ uniformly pushes the patch embeddings away from each other so that the representation space does not collapse.

\begin{figure*}[t!]
\begin{center}
\includegraphics[width=\textwidth]{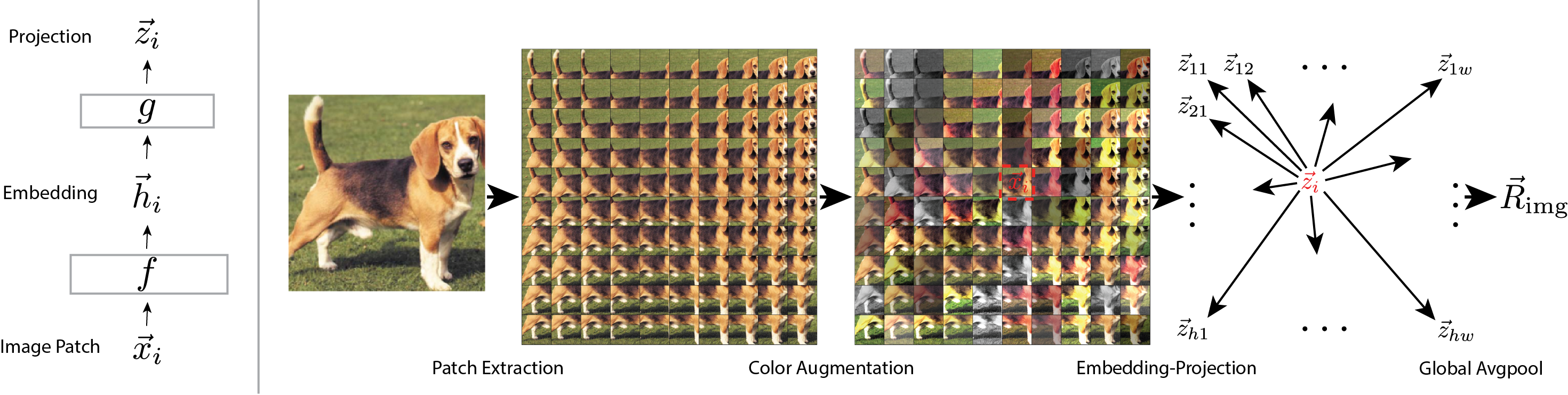}
\vspace{-0.25in}
\caption{\textbf{BagSSL pipeline.} From each image, fixed-size image patches are extracted, color-augmented, encoded to embedding and projection space. During training, patch projections of co-occurring patches (from the same image) are pulled together while an anti-collapse regularization is applied to push non co-occurring patch projections away. After training, patch embeddings $\{\vec{h}\}$ from the same image are averaged to form the image representation $\vec{R}_{img}$.}
\label{fig:drip_figure}
\end{center}
\end{figure*}

{\bf Bag-of-Local-Feature Evaluation.} After we have learned a representation of fix-scale image patches, we can embed all of the image patches $\{\vec{x}_{11},\dots,\vec{x}_{HW}\}$ within an image into the embedding space. Then, we can obtain the representation $R_{img}$ for the whole image by linearly aggregating (averaging) all patch embeddings. The pipeline is shown in Figure~\ref{fig:drip_figure}. $\vec{z}=g(\vec{h};\psi)$ and $\vec{h} = f(\vec{x};\theta)$. We call $\vec{h}$ the {\it embedding} and $\vec{z}$ the {\it projection} of an image patch, $\vec{x}$. $\{\vec{x}\}$ are fixed-scale. The function $f(\cdot;\theta)$ is a deep neural network with parameters $\theta$, and $g$ is typically a much simpler neural network with only one or a few fully connected layers and parameters $\psi$. Following a convention in joint-embedding SSL, we use $\vec{h}$ as patch representation rather than $\vec{z}$ for slightly better accuracy, i.e., $\vec{R}_{img} = \text{mean}_{HW}(\vec{h}_{11},\cdots,\vec{h}_{HW})$. During SSL training, given an image patch $\vec{x}_i$, the objective tries to pull its projection $\vec{z}_i$ closer to the projections of other co-occurring image patches and to push away non-co-occurring image patches' projections. For easier reference, we denote this process as {\it BagSSL}.

\section{Empirical Results}
\label{sec:experiments}
In this section, we present a series of empirical results and test BagSSL with several representative joint-embedding SSL methods (SimCLR\citep{chen2020simple}, VICReg\citep{bardes2021vicreg}, TCR\citep{li2022neural}, and BYOL\citep{grill2020BYOL}) on four standard benchmarks (CIFAR10, CIFAR100, ImageNet-100, and ImageNet-1K). Through experiments, we have the following observations:
\begin{itemize}
    \item BagSSL achieves similar or even better results than baseline methods. Even with $32\times 32$ patch representation, BagSSL achieves $62\%$ top-1 linear probing accuracy on ImageNet.
    \item The aggregated patch embedding converges to the whole-image representation in baseline SSL methods with respect to the number of patches aggregated.
    \item Through visualization, we show that the smaller-scale patch representation preserves locality better.
    \item We can leverage BagSSL to further improve representation quality of the baseline methods.
\end{itemize}

The first two observations show from two different aspects that a patch-based SSL representation explains the performance. The third observation shows how patch representation supplements the prevailing invariance perspective. And the fourth observation provides practical benefit in addition to understanding.

\subsection{BagSSL Versus Baseline Methods}
\label{sec:bagssl_baseline}
\begin{figure}[t]
\begin{center}
\includegraphics[width=\textwidth]{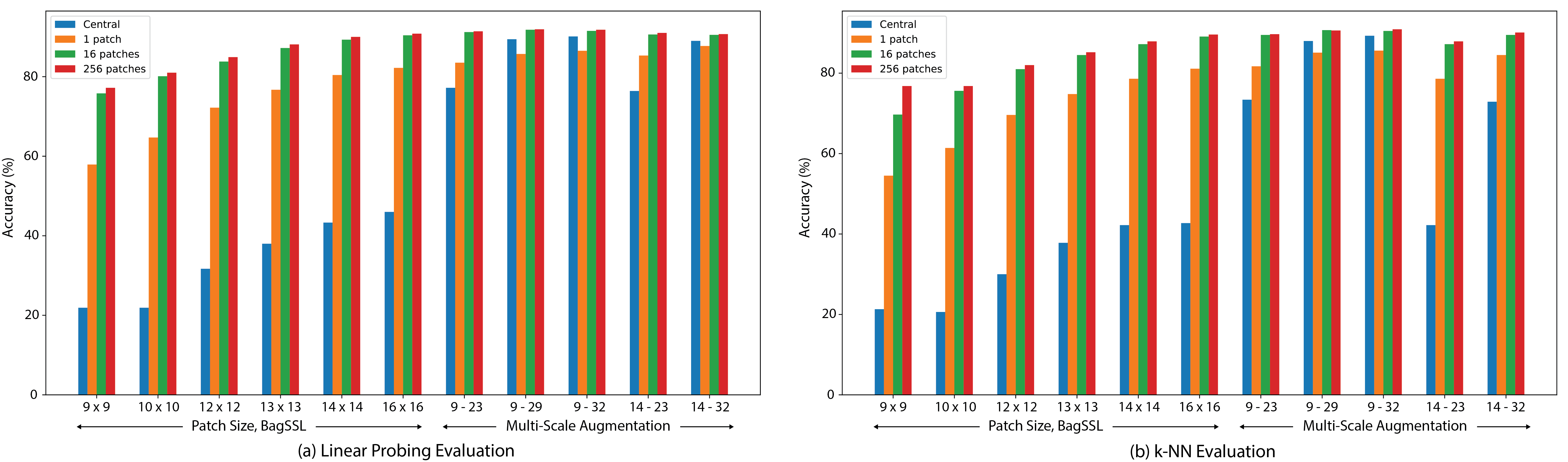}
\vspace{-0.3in}
\caption{\textbf{Linear probing and kNN evaluation on CIFAR-10 are consistent.} We evaluate the performance of a linear classifier (a) and a k-NN classifier (b) for pretraining with various patch sizes and various evaluation setups. During pretraining, BagSSL uses fixed-scale patches, and baseline methods use multi-scale augmentation \code{RandomResizedCrop(min\_scale, max\_scale)}. For example, \code{RandomResizedCrop(0.08, 1.0)} corresponds to randomly and uniformly select crops ranging from $9\times 9$ to $32\times 32$. The patch size and augmentation crop size range are marked in the figure.
The ``Central'' evaluation is the standard evaluation protocol where the classifier is trained and evaluated on single fixed central patches of the image, which is the entire image for CIFAR-10. For the $n$ patch evaluation, the classifier is trained and evaluated on the linearly-aggregated embedding of $n$ patches, sampled with the same scale factor as during pretraining. Please note that the ``central'' evaluation is expected to perform poorly on fix-scale pretraining as the model has never seen the entire image during pretraining.}
\label{fig:cifar_linear_knn}
\end{center}
\vspace{-0.2in}
\end{figure}

\begin{figure}[t]
\begin{center}
\includegraphics[trim=0.15cm 0.0cm 0.15cm 0cm,width=\textwidth]{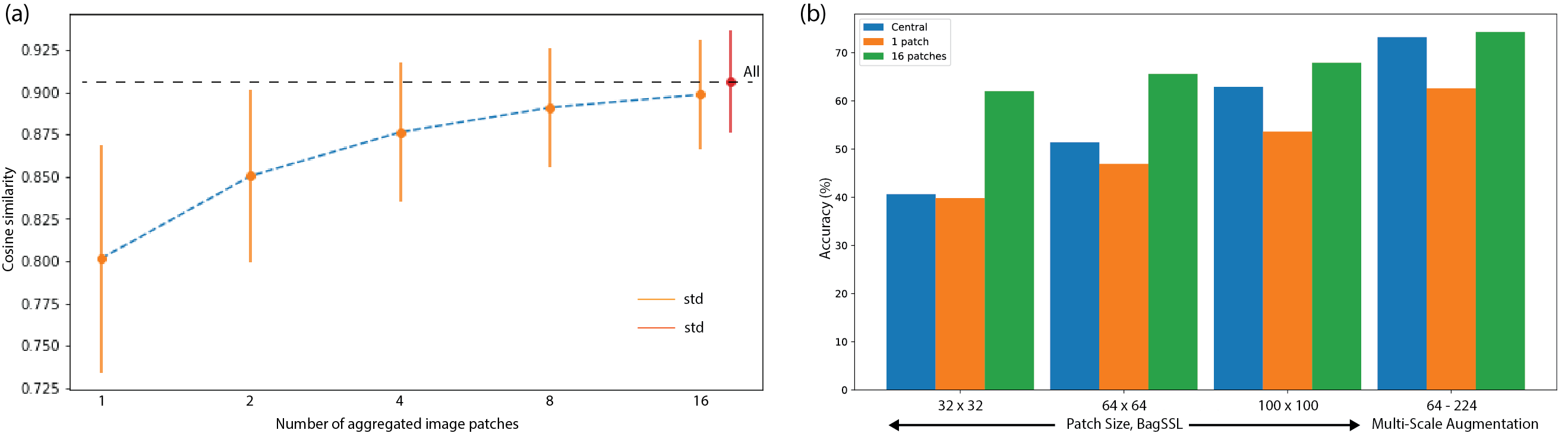}
\vspace{-0.1in}
\caption{\textbf{(a) Patch embedding convergence to the instance embedding.} For a baseline multi-scale pretrained VICReg model, we show that the patch embedding aggregation converges to the whole-image embedding as the number of aggregated patches increases. We evaluate the cosine similarity between the aggregation of $N$ patch embeddings and the whole-image embedding. $N$ is selected from $1, 2, 4, 8, 16$ and all possible patches in the image. \textbf{(b) Linear evaluation on ImageNet for various \code{RandomResizedCrop} scales.}  We show the performance of a linear classifier for various pretraining and evaluation settings. BagSSL uses fixed-scale patches, and the baseline method, VICReg, uses multi-scale augmentation \code{RandomResizedCrop(0.08, 1.0)}, which corresponds to randomly and uniformly select crop scale ranging from $64\times 64$ to $224\times 224$. For each pretraining setting, we provide three different evaluations: 1) ``Central'': the standard evaluation, which takes a $224\times 224$ crop; 2) ``$1$ patch'': this evaluation takes $1$ patch or crop following the corresponding pretraining setting; 3) ``$16$ patches'': this evaluation takes $16$ randomly selected patches or crops following the corresponding pretraining setting.}
\label{fig:drip_imagenet_linear_800ep}
\end{center}
\end{figure}

We first test BagSSL with four baseline methods. The only difference between BagSSL and the baselines during training is that we replace multi-scale cropping augmentation with fixed-scale image patches. Similarly, two image patches are randomly selected from each image in the batch. While we mainly present linear probing accuracy, Figure~\ref{fig:cifar_linear_knn} shows that k-NN evaluation is consistent with linear probing evaluation. During the evaluation, we show standard central crop evaluation results for both BagSSL and baselines. Further, we show the multi-patch aggregated evaluation and multi-crop aggregated evaluation. In the tables, ``patch'' means that fixed-scale patch embeddings are aggregated, and ``crop'' means that multi-scale crop embeddings are aggregated. All implementation details can be found in Appendix \ref{appendix: Details}. The main results are highlighted in Table~\ref{tab:cifar10_other_methods}, Table~\ref{tab:cifar100_other_methods}, Table~\ref{tab:IN100_other_methods}, where BagSSL matches the baselines or surpass them on CIFAR10, CIFAR100, and ImageNet-100 datasets. In Figure~\ref{fig:drip_imagenet_linear_800ep}, we show the results on ImageNet-1K. Interestingly, with $32\times 32$ patch representation, BagSSL achieves $62\%$ top-1 linear probing accuracy. There is a $5.3\%$ performance gap between BagSSL and VICReg on ImageNet-1K, possibly due to the engineering issue.

{\bf CIFAR.} We first provide experimental results on the standard CIFAR-10 and CIFAR-100 datasets \citep{krizhevsky2020cifar} using ResNet-34. During training, fixed-scale image patches are used for BagSSL, and baseline methods RandomResizedCrop(0.08, 1.0) augmentation, which means that we randomly choose to sample from $9\times 9$ to $32 \times 32$ patches as the crops. Sampled patches and crops are upsampled to a uniform $32\times 32$ resolution before embedding by ResNet-34. The standard evaluation method generates the embedding using the entire image, both during training of the linear classifier and at final evaluation, and it is denoted as (\textit{Central}). An alternative evaluation is that an image embedding is generated by inputting a certain number of patches (same scale as training time and upsampled) into the neural network and aggregating the patch embeddings by averaging. The number of aggregated patches is marked in the Figures and Tables. In Figure~\ref{fig:cifar_linear_knn}, we show that linear probing and k-NN evaluations are consistent. In the following, we mainly present linear probing evaluation. 
The results on CIFAR-10 and CIFAR-100 are shown in Tables~\ref{tab:cifar10_other_methods} and \ref{tab:cifar100_other_methods} respectively. The main observation is that pretraining on small patches and evaluating with the averaged embedding performs on par or better than the baseline methods, as highlighted. The central crop evaluation performs significantly worse for patch-based training is expected since the network has never seen a whole-image during training.

\begin{table}[t]
\caption{\textbf{Performance on CIFAR-10 with patch-based and standard multi-Scale SSL pre-training.} We evaluate the performance of a linear classifier with BagSSL and baseline SSL methods.  BagSSL uses \textit{patch-based training}, where $14\times 14$ image patches are sampled during pretraining.  Baseline methods use \textit{multi-scale training}, where the patch scale is uniformly sampled between scale 0.08 and 1.0, which means that multi-scale crops are sampled uniformly from $9\times 9$ to $32 \times 32$ patches.}
\centering
\vspace{2mm}
\label{tab:cifar10_other_methods}

\resizebox{\columnwidth}{!}{%
\begin{tabular}{lcccccccc}
\toprule

Method & \multicolumn{4}{c}{\em Patch-based training evaluation} & \multicolumn{4}{c}{\em Multi-scale training evaluation} \\
 & Central crop   & 1 patch    & 16 patches & 256 patches & Central crop   & 1 crop    & 16 crops & 256 crops \\
\midrule
SimCLR          &  46.2 & 82.1  &  90.5 & \cellcolor{pearDark!20}90.8 & \cellcolor{pearDark!20}90.2 & 86.4 & 91.6 & 91.8 \\
TCR             &  46.0 & 82.2  &  90.4 & \cellcolor{pearDark!20}90.8 & \cellcolor{pearDark!20}90.1 & 86.5 & 91.5 & 91.8 \\
VICReg          &  47.1 & 83.1  &  90.9 & \cellcolor{pearDark!20}91.2 & \cellcolor{pearDark!20}90.7 & 87.3 & 91.9 & 92.0 \\
BYOL            &  47.3 & 83.6  &  91.3 & \cellcolor{pearDark!20}91.5 & \cellcolor{pearDark!20}90.9 & 87.8 & 92.3 & 92.4 \\
\bottomrule
\end{tabular}
}
\end{table}

\begin{table}[t]
\caption{\textbf{Performance on CIFAR-100 with patch-based and standard multi-scale SSL pre-training.} We evaluate the performance of a linear classifier with BagSSL and baseline SSL methods.  BagSSL uses \textit{patch-based training}, where $14\times 14$ image patches are sampled during pretraining.  Baseline methods use \textit{multi-scale training}, where the patch scale is uniformly sampled between scale 0.08 and 1.0, which means that multi-scale crops are sampled uniformly from $9\times 9$ to $32 \times 32$ patches.}
\centering
\vspace{2mm}
\label{tab:cifar100_other_methods}

\resizebox{\columnwidth}{!}{%
\begin{tabular}{lcccccccc}
\toprule

Method & \multicolumn{4}{c}{\em Patch-based training evaluation} & \multicolumn{4}{c}{\em Multi-scale training evaluation} \\
 & Central crop   & 1 patch    & 16 patches & 256 patches & Central crop   & 1 crop   & 16 crops & 256 crops \\
\midrule
SimCLR                  &  34.7 & 59.4  & 67.2 & \cellcolor{pearDark!20}67.4 & \cellcolor{pearDark!20}66.8 & 60.5 & 68.2 & 68.3 \\
TCR                     &  34.6 & 59.2  & 67.1 & \cellcolor{pearDark!20}67.3 & \cellcolor{pearDark!20}66.8 & 60.5 & 68.1 & 68.3 \\
VICReg          &  35.5 & 60.1  & 68.0 & \cellcolor{pearDark!20}68.3 & \cellcolor{pearDark!20}67.6 & 61.4 & 69.0 & 69.3 \\
BYOL             &  37.4 & 60.9  & 68.9 & \cellcolor{pearDark!20}69.2 & \cellcolor{pearDark!20}68.8 & 62.3 & 69.7 & 69.9 \\
\bottomrule
\end{tabular}
}
\end{table}

{\bf ImageNet.} We also provide experimental results on ImageNet-100 and ImageNet datasets \cite{deng2009imagenet} with ResNet-50 and linear probing evaluation protocol. ImageNet-100 and ImageNet results are shown in Table~\ref{tab:IN100_other_methods} and Figure~\ref{fig:drip_imagenet_linear_800ep}(b) respectively. The behavior observed on CIFAR-10 generalizes to ImageNet-100. Averaging embeddings of small patches produced by the patch-based pretrained models perform comparably to standard ``central'' evaluation of the embedding produced by the baseline models, as highlighted in Table~\ref{tab:IN100_other_methods}. In Figure~\ref{fig:drip_imagenet_linear_800ep}(b), we show the results on ImageNet. The patch-based pretrained model achieves $67.9\%$ top-1 accuracy with $16$ patch embedding averaged, whereas the baseline method VICReg achieves $73.2\%$ top-1 accuracy. This $5.3\%$ performance gap might be due to sub-optimal hyperparameters as we did not optimize the hyperparameters for patch-based training pretraining. Interestingly, with $32\times 32$ patch representation, BagSSL achieves $62\%$ top-1 linear probing accuracy.

\begin{table}[t]
\caption{\textbf{Performance on ImageNet-100 with patch-based and standard multi-scale SSL pre-training.} We evaluate the performance of a linear classifier with BagSSL and baseline SSL methods. BagSSL uses \textit{patch-based training}, where $100\times 100$ patches are sampled during pretraining. Baseline methods use \textit{multi-scale training}, where the patch scale is uniformly sampled between scale 0.08 and 1.0, which means that multi-scale crops are sampled uniformly from $64\times 64$ to $224 \times 224$ patches.}
\centering
\vspace{2mm}
\label{tab:IN100_other_methods}

\resizebox{\columnwidth}{!}{%
\begin{tabular}{lcccccccc}
\toprule

Method & \multicolumn{4}{c}{\em Patch-based training evaluation} & \multicolumn{4}{c}{\em Multi-scale training evaluation} \\
 & Central crop   & 1 patch    & 16 patches & 48 patches & Central crop   & 1 crop   & 16 crops & 48 crops \\
\midrule
SimCLR      & 41.4  &  45.7 & 76.1 & \cellcolor{pearDark!20}76.2 & \cellcolor{pearDark!20}77.5 & 70.3 & 78.6 & 79.0 \\
TCR         & 41.3  &  45.6 & 76.1 & \cellcolor{pearDark!20}76.3 & \cellcolor{pearDark!20}77.3 & 70.1 & 78.5 & 78.8 \\
VICReg      & 42.1  &  46.1 & 76.8 & \cellcolor{pearDark!20}76.9 & \cellcolor{pearDark!20}77.8 & 70.7 & 79.1 & 79.4 \\
BYOL        & 42.9  &  47.3 & 77.9 & \cellcolor{pearDark!20}77.7 & \cellcolor{pearDark!20}78.0 & 71.1 & 79.4 & 80.1 \\
\bottomrule
\end{tabular}
}
\end{table}

\subsection{Patch Embedding Aggregation Converges to the Whole-Image Embedding}
In this experiment, we show that for a multi-scale pretrained baseline SSL model, VICReg, linearly aggregating the patch embedding converges to the whole-image embedding. This baseline VICReg network checkpoint has been trained on image crops ranging from $64\times 64$ to $224\times 224$. We randomly select 512 images from the ImageNet dataset. For each image, we first get the $224\times 224$ center crop embedding. We also average embeddings of $N$ random $100\times 100$ patches for each image. Then we calculate the cosine similarity between the patch-aggregated and center-crop embedding. Figure~\ref{fig:drip_imagenet_linear_800ep}(a) shows that the aggregated embedding converges to the whole-image embedding as $N$ increases from $1$ to $16$ to all the image patches\footnote{``All'': extracting overlapped patches with stride $4$ and totally aggregate about 1000 patches' embeddings per image.}.

\subsection{Patch Representation Preserves Locality}
\label{sec:visualization}

\begin{figure}[t]
\begin{center}
\includegraphics[trim=0.13cm 0.0cm 0.2cm 0cm,width=\textwidth]{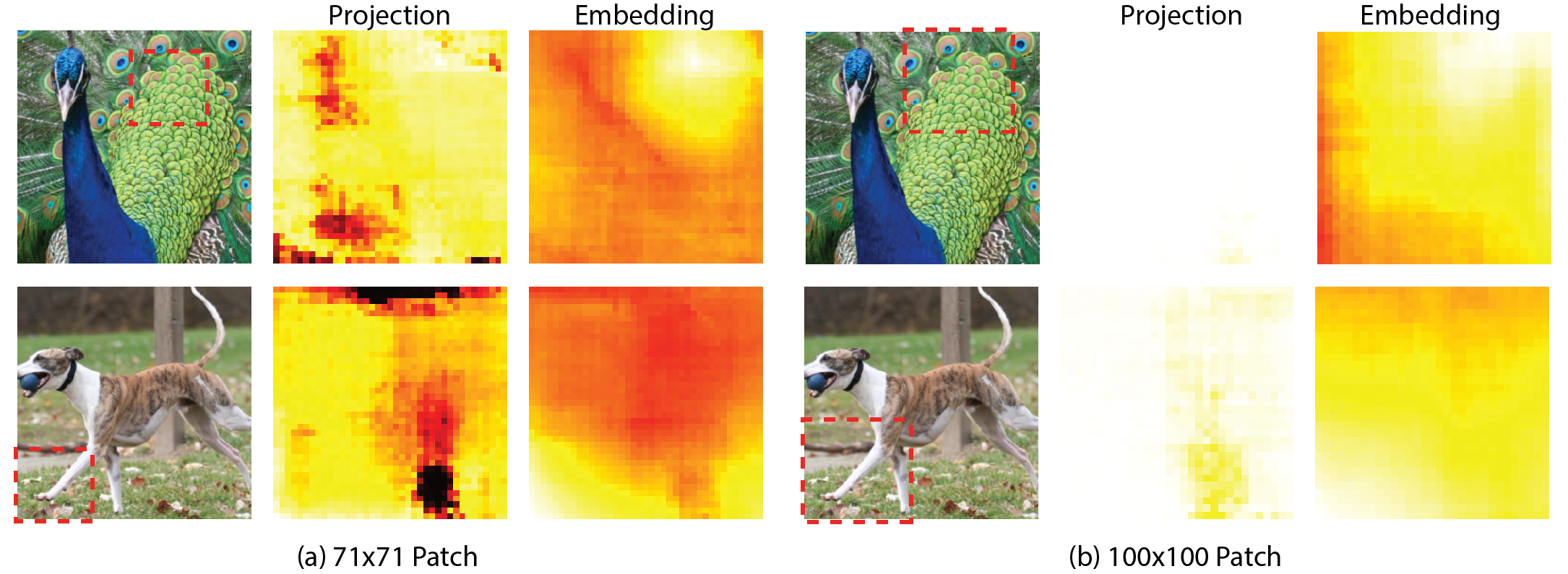}
\vspace{-1em}
\caption{\textbf{Visualization of cosine similarity in the projection space and the embedding space.} Query patch is indicated by red dash. Projection and Embedding cosine-similarity heatmaps use the same color scaling. The projection vectors are significantly more invariant compared to the embedding ones, and the embedding space contains localized information that is shared among similar patches, when the size of the patches is small enough. We can see that the embedding space tends to preserve more locality compared to the projection space.}
\label{fig:IN_heatmap_main}
\end{center}
\vspace{-0.2in}
\end{figure}

\begin{figure}[t]
\begin{center}
\includegraphics[width=\textwidth]{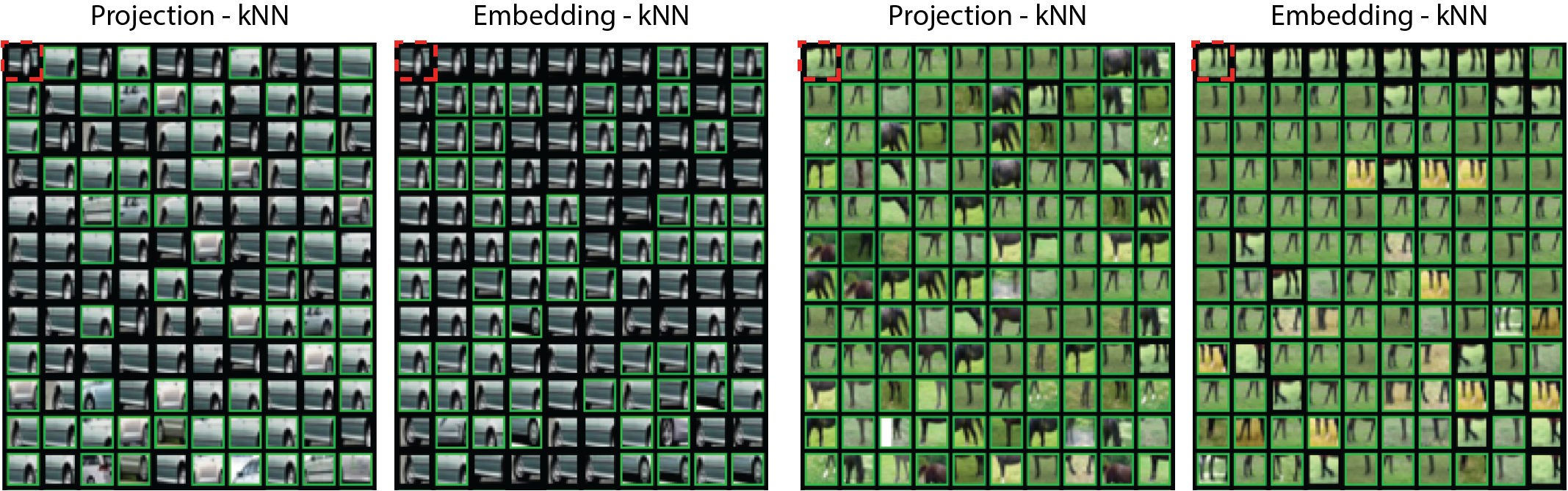}
\vspace{-1em}
\caption{\textbf{Visualization of kNN in the projection space and the embedding space for CIFAR10.} Distance is calculated by cosine similarity. Query patch is in the top left corner encircled by red-dash box, green box indicates patches from other image of the same class. Patches without surrounding box is from the same image as the query. While the nearest neighbors are both from same-category instances, we can see that the embedding space tends to preserve the local part information, whereas the projection space may collapse different parts of the same category.}
\label{fig:cifar_knn_main}
\end{center}
\vspace{-0.2in}
\end{figure}

Joint-embedding SSL methods are primarily motivated from an invariance perspective. While this perspective is relatively accurate globally, we show that locality is preserved when we zoom into local patch-level representation. In this section, we provide nearest neighbor visualization on CIFAR-10 and cosine-similarity heatmap visualization on ImageNet to show this point and provide a deeper understanding of the learned representation.

First, we take a baseline VICReg model pretrained on ImageNet, for a given image patch (e.g., circled by red dash boxes in Figure~\ref{fig:IN_heatmap_main}), we visualize the cosine-similarity between embedding from this patch and that from the other same-size patches from the same image. The heatmap visualization is normalized to the same scale. While the cosine-similarity heatmap for large ($100\times 100$) patches is more invariant, as shown in Figure~\ref{fig:IN_heatmap_main}(b), the heatmap for smaller ($71\times 71$) patches preserves the locality relatively well as demonstrated in Figure~\ref{fig:IN_heatmap_main}(a).

Next, we pre-train a model with only $14\times 14$ image patches on CIFAR-10 and calculate the projection and embedding vectors of all different image patches from the training set. Then for a given $14\times 14$ image patch (e.g., the ones circled by red dash boxes Fig~\ref{fig:cifar_knn_main}), we visualize its $k$ nearest neighbors in terms of cosine-similarity in both the projection and the embedding space. Figure~\ref{fig:cifar_knn_main} shows the results for two image patches. The patches circled by green boxes are image patches from another image of the same category, whereas the uncircled patches are from the same image.

Overall, we observe that the projection vectors are significantly more invariant than the embedding vectors. This is apparent from Figure \ref{fig:IN_heatmap_main} and Figure~\ref{fig:cifar_knn_main}. For the CIFAR kNN patches, NNs in the embedding space are visually much more similar than NNs in the projection space. In the embedding space, the nearest NNs are mostly locally shifted patches of similar ``part'' information. For projection space, however, many NNs are patches of different ``part'' information from the same class. E.g., we can see in Figure~\ref{fig:cifar_knn_main} that an NNs of a ``wheel'' in the projection space might be a ``door'' or a ``window''. However, the NNs in the embedding space all contain ``wheel'' information. In the second example, the NNs of a ``horse legs'' patch may have different ``horse'' body parts, whereas the NNs in the embedding space are all ``horse leg''.

The heatmap visualization on ImageNet also illustrates the same phenomenon. The projection vector from a patch is highly similar to that from the query patch whenever the patch has enough information to infer the class of the image. While for embedding vectors, the similarity area is much more localized to the query patch or other patches with similar features (the other leg of the dog in Figure \ref{fig:IN_heatmap_main}). This general observation is consistent with the results of the visualizations in \citet{bordes2021HDvisualizationofSSLrepre}. A more thorough visualization is provided in Appendix~\ref{app:visualization}.

\subsection{Patch-Based Evaluation Enhances the SSL Baselines}

{\bf Global Aggregation.} The results in Section~\ref{sec:bagssl_baseline} show that the best performance is obtained when the pretraining step uses multi-scale crops and the evaluation step uses the aggregated patch embeddings. Here, we evaluate the patch-aggregated representation with four baseline methods pretrained on ImageNet to further confirm this performance boost. All models are downloaded from their original repository. Table \ref{tab:drip_other_methods} shows the linear evaluation performance on the validation set of ImageNet using the whole-image embedding and patch-aggregated embedding. For all these models, aggregated embedding outperforms whole-image evaluation, often by more than 1\%. Also, increasing the number of patches averaged in the aggregation process improves the performance. We do not go beyond 48 patches due to engineering limits on memory and runtime. Still, we hypothesize that a further increase in the number of patches will improve the performance, as demonstrated on CIFAR-10, where 256 patches significantly outperform 16 patches.

\begin{table}[t]
\caption{\textbf{Linear evaluation with aggregated embedding on ImageNet with models trained with state-of-the-art SSL methods.} Using aggregated embedding outperforms embedding from the center crop. Central: embedding from the center cropped image is used in training and testing using the standard linear evaluation protocol. 1, 16, and 48 patches: The linear classifier is trained and evaluated on the aggregated embedding of 1, 16, and 48 patches respectively, sampled with the same scale factor range as during pretraining (0.08, 1.0).}
\centering
\vspace{2mm}
\label{tab:drip_other_methods}
\begin{tabular}{lcccc}
\toprule
Method          & Central   & 1 patch    & 16 patches & 48 patches \\
\midrule
SimCLR          & 69.3      & 54.3     & 71.0     &  71.3 \\
TCR             & 69.0      & 54.1     & 70.7     &  71.1 \\
VICReg          & 73.2      & 57.6     & 74.2     &  74.4 \\
BYOL            & 74.3      & 59.3     & 75.4     &  75.6 \\
\bottomrule
\end{tabular}
\end{table}


{\bf Local Aggregation.}
In this work, the embedding aggregation is global or uses bag-of-local features. The drawback of doing so is that spatial information can be lost. An alternative way to aggregate patch embeddings is local averaging and concatenating locally averaged embeddings into a single long vector. Empirically, such a method tends to outperform the global average significantly. To show the results, we use the checkpoints of the SOTA SSL model pretrained on the CIFAR10 dataset from SoloLearn \citep{guilherme2022sololearn} and tested linear and kNN accuracy with local concatenation aggregation. As shown in Table~\ref{tab:sota}, concatenation aggregation further improve the performance of these SOTA SSL model. Even with only 25 patches, the k-NN accuracy of the aggregated embedding outperforms the baseline linear evaluation accuracy by a larger margin compared to global aggregation results shown in Table~\ref{tab:cifar10_other_methods}.

\begin{table}
\caption{ \textbf{Evaluation of SOTA SSL models and these models with linearly-aggregated patches embedding enhancement.} All the baseline SSL model uses ResNet-18 as the backbone. We apply spatial average pooling on the last layer output of ResNet-18 and treat it as feature. We evaluate the performance of these checkpoints with both linear classifier and K-nearest-neighbor (KNN) classifier. For the ``Enhancement'' evaluation, the KNN classifier is evaluated on the linearly-aggregated embedding of 25 patches with size $16 \times 16$. These patches are sampled using a sliding window with stride $4$.}
\centering
\vspace{2mm}
\label{tab:sota}
\begin{tabular}{lccc}
\toprule
Method          & {\em Baseline (KNN)}   & {\em Baseline (linear)}    & {\em Local Aggregation (KNN)} \\
\midrule
SimCLR           & 90.2     & 90.7     & 93.1    \\
VICReg           & 90.8      & 91.2     & 93.1    \\
BYOL             & 91.5      & 92.6     & 93.5    \\
\bottomrule
\end{tabular}
\end{table}

\section{Discussion}
In this paper, we seek to understand the success of joint-embedding SSL methods. We establish a formal connection between joint-embedding SSL and the co-occurrence of image patches. We demonstrate learning an embedding for fixed-size image patches and linear aggregating the local patch embeddings can achieve similar or even better performance than the baseline models pretrained with multi-scale crops. On the other hand, with a multi-scale pretrained model, we show that the whole image embedding is approximately the average of local patch embeddings. Given this, invariance is expected at the global scale. Through visualization, we show that the locality is preserved when we zoom into local patch-level representation. These findings supplement the prevailing invariance perspective and show that there is a distributed representation of local image patches behind the success of joint-embedding SSL. The insights from this angle also help us enhance the representation quality from the baseline methods.

There are a few limitations to this work. First, we still use the commonly used projector head, whose role is yet to be fully understood. Second, we adopted the convention of two-crop training for BagSSL. If joint-embedding SSL aims to model the co-occurrence of patches, two local patches would be relatively inefficient and create a disadvantage during training. Third, we haven't optimized the hyperparameters thoroughly for BagSSL on ImageNet-1K due to engineering resource limitations. Given the second and third limitations, BagSSL results may be further improved, given better engineering. We leave these to our future work.

\subsubsection*{Acknowledgments}
We thank Zeyu Yun for helping us with the local aggregation evaluation during the preparation of this paper. We thank Pierre-Etienne Fiquet, Eero Simoncelli, Thomas Yerxa, and many other friends at Flatiron Institute for their valuable comments and suggestions.

\bibliography{iclr2023_conference}

\begin{thebibliography}{37}
\providecommand{\natexlab}[1]{#1}
\providecommand{\url}[1]{\texttt{#1}}
\expandafter\ifx\csname urlstyle\endcsname\relax
  \providecommand{\doi}[1]{doi: #1}\else
  \providecommand{\doi}{doi: \begingroup \urlstyle{rm}\Url}\fi

\bibitem[Bao et~al.()Bao, Dong, Piao, and Wei]{bao2021ImageBert}
Hangbo Bao, Li~Dong, Songhao Piao, and Furu Wei.
\newblock Beit: {BERT} pre-training of image transformers.
\newblock In \emph{The International Conference on Learning Representations,
  {ICLR} 2022}.

\bibitem[Bardes et~al.(2021)Bardes, Ponce, and LeCun]{bardes2021vicreg}
Adrien Bardes, Jean Ponce, and Yann LeCun.
\newblock Vicreg: Variance-invariance-covariance regularization for
  self-supervised learning.
\newblock In \emph{International Conference on Learning Representations,
  {ICLR}}, 2021.

\bibitem[Bardes et~al.(2022)Bardes, Ponce, and LeCun]{bardes2022vicregl}
Adrien Bardes, Jean Ponce, and Yann LeCun.
\newblock Vicregl: Self-supervised learning of local visual features.
\newblock In \emph{Advances in Neural Information Processing Systems,
  {NeurIPS}}, 2022.

\bibitem[Bordes et~al.(2022)Bordes, Balestriero, and
  Vincent]{bordes2021HDvisualizationofSSLrepre}
Florian Bordes, Randall Balestriero, and Pascal Vincent.
\newblock High fidelity visualization of what your self-supervised
  representation knows about.
\newblock \emph{Transactions on Machine Learning Research, {TMLR}}, 2022.

\bibitem[Brendel \& Bethge(2018)Brendel and Bethge]{brendel2018BagNet}
Wieland Brendel and Matthias Bethge.
\newblock Approximating cnns with bag-of-local-features models works
  surprisingly well on imagenet.
\newblock In \emph{International Conference on Learning Representations,
  {ICLR}}, 2018.

\bibitem[Bromley et~al.(1993)Bromley, Guyon, LeCun, S{\"a}ckinger, and
  Shah]{bromley1993signature}
Jane Bromley, Isabelle Guyon, Yann LeCun, Eduard S{\"a}ckinger, and Roopak
  Shah.
\newblock Signature verification using a" siamese" time delay neural network.
\newblock \emph{Advances in Neural Information Processing Systems, {NeurIPS}},
  1993.

\bibitem[Chen et~al.(2020{\natexlab{a}})Chen, Kornblith, Norouzi, and
  Hinton]{chen2020simple}
Ting Chen, Simon Kornblith, Mohammad Norouzi, and Geoffrey Hinton.
\newblock A simple framework for contrastive learning of visual
  representations.
\newblock In \emph{International Conference on Machine Learning, {ICML}},
  2020{\natexlab{a}}.

\bibitem[Chen \& He(2021)Chen and He]{chen2021simsiam}
Xinlei Chen and Kaiming He.
\newblock Exploring simple siamese representation learning.
\newblock In \emph{Proceedings of the Conference on Computer Vision and Pattern
  Recognition, {CVPR}}, 2021.

\bibitem[Chen et~al.(2020{\natexlab{b}})Chen, Fan, Girshick, and
  He]{chen2020improved}
Xinlei Chen, Haoqi Fan, Ross Girshick, and Kaiming He.
\newblock Improved baselines with momentum contrastive learning.
\newblock \emph{arXiv preprint arXiv:2003.04297}, 2020{\natexlab{b}}.

\bibitem[Chen et~al.(2018)Chen, Paiton, and Olshausen]{chen2018sparse}
Yubei Chen, Dylan Paiton, and Bruno Olshausen.
\newblock The sparse manifold transform.
\newblock \emph{Advances in Neural Information Processing Systems, {NeurIPS}},
  2018.

\bibitem[da~Costa et~al.(2022)da~Costa, Fini, Nabi, Sebe, and
  Ricci]{guilherme2022sololearn}
Victor Guilherme~Turrisi da~Costa, Enrico Fini, Moin Nabi, Nicu Sebe, and Elisa
  Ricci.
\newblock solo-learn: A library of self-supervised methods for visual
  representation learning.
\newblock \emph{Journal of Machine Learning Research, {JMLR}}, 2022.

\bibitem[Deng et~al.(2009)Deng, Dong, Socher, Li, Li, and
  Fei-Fei]{deng2009imagenet}
Jia Deng, Wei Dong, Richard Socher, Li-Jia Li, Kai Li, and Li~Fei-Fei.
\newblock Imagenet: A large-scale hierarchical image database.
\newblock In \emph{Proceedings of the Conference on Computer Vision and Pattern
  Recognition, {CVPR}}, 2009.

\bibitem[Dosovitskiy et~al.(2016)Dosovitskiy, Fischer, Springenberg,
  Riedmiller, and Brox]{DosovitskiyFSRB16}
Alexey Dosovitskiy, Philipp Fischer, Jost~Tobias Springenberg, Martin~A.
  Riedmiller, and Thomas Brox.
\newblock Discriminative unsupervised feature learning with exemplar
  convolutional neural networks.
\newblock \emph{IEEE Transactions on Pattern Analysis and Machine Intelligence,
  {PAMI}}, 2016.

\bibitem[Dosovitskiy et~al.(2021)Dosovitskiy, Beyer, Kolesnikov, Weissenborn,
  Zhai, Unterthiner, Dehghani, Minderer, Heigold, Gelly, Uszkoreit, and
  Houlsby]{dosovitskiy2020image}
Alexey Dosovitskiy, Lucas Beyer, Alexander Kolesnikov, Dirk Weissenborn,
  Xiaohua Zhai, Thomas Unterthiner, Mostafa Dehghani, Matthias Minderer, Georg
  Heigold, Sylvain Gelly, Jakob Uszkoreit, and Neil Houlsby.
\newblock An image is worth 16x16 words: Transformers for image recognition at
  scale.
\newblock In \emph{The International Conference on Learning Representations,
  {ICLR}}, 2021.

\bibitem[Dumais(2004)]{dumais2004latent}
Susan~T Dumais.
\newblock Latent semantic analysis.
\newblock \emph{Annual Review of Information Science and Technology, {ARIST}},
  2004.

\bibitem[Ermolov et~al.(2021)Ermolov, Siarohin, Sangineto, and
  Sebe]{ermolov2021whitening}
Aleksandr Ermolov, Aliaksandr Siarohin, Enver Sangineto, and Nicu Sebe.
\newblock Whitening for self-supervised representation learning.
\newblock In \emph{International Conference on Machine Learning, {ICML}}, 2021.

\bibitem[Garrido et~al.(2023)Garrido, Chen, Bardes, Najman, and
  Lecun]{Garrido2023duality}
Quentin Garrido, Yubei Chen, Adrien Bardes, Laurent Najman, and Yann Lecun.
\newblock On the duality between contrastive and non-contrastive
  self-supervised learning.
\newblock \emph{The International Conference on Learning Representations,
  {ICLR}}, 2023.

\bibitem[Gidaris et~al.(2020)Gidaris, Bursuc, Komodakis, P{\'e}rez, and
  Cord]{gidaris2020SSLbypredBOW}
Spyros Gidaris, Andrei Bursuc, Nikos Komodakis, Patrick P{\'e}rez, and Matthieu
  Cord.
\newblock Learning representations by predicting bags of visual words.
\newblock In \emph{Proceedings of the Conference on Computer Vision and Pattern
  Recognition, {CVPR}}, 2020.

\bibitem[Grill et~al.(2020)Grill, Strub, Altch{\'{e}}, Tallec, Richemond,
  Buchatskaya, Doersch, Pires, Guo, Azar, Piot, Kavukcuoglu, Munos, and
  Valko]{grill2020BYOL}
Jean{-}Bastien Grill, Florian Strub, Florent Altch{\'{e}}, Corentin Tallec,
  Pierre~H. Richemond, Elena Buchatskaya, Carl Doersch, Bernardo~{\'{A}}vila
  Pires, Zhaohan Guo, Mohammad~Gheshlaghi Azar, Bilal Piot, Koray Kavukcuoglu,
  R{\'{e}}mi Munos, and Michal Valko.
\newblock Bootstrap your own latent - {A} new approach to self-supervised
  learning.
\newblock In \emph{Advances in Neural Information Processing Systems,
  {NeurIPS}}, 2020.

\bibitem[HaoChen et~al.(2021)HaoChen, Wei, Gaidon, and Ma]{haochen2021provable}
Jeff~Z HaoChen, Colin Wei, Adrien Gaidon, and Tengyu Ma.
\newblock Provable guarantees for self-supervised deep learning with spectral
  contrastive loss.
\newblock \emph{Advances in Neural Information Processing Systems, {NeurIPS}},
  2021.

\bibitem[He et~al.(2020)He, Fan, Wu, Xie, and Girshick]{he2020MoCo}
Kaiming He, Haoqi Fan, Yuxin Wu, Saining Xie, and Ross Girshick.
\newblock Momentum contrast for unsupervised visual representation learning.
\newblock In \emph{Proceedings of the Conference on Computer Vision and Pattern
  Recognition, {CVPR}}, 2020.

\bibitem[He et~al.(2022)He, Chen, Xie, Li, Doll{\'{a}}r, and
  Girshick]{he2021MAESSL}
Kaiming He, Xinlei Chen, Saining Xie, Yanghao Li, Piotr Doll{\'{a}}r, and
  Ross~B. Girshick.
\newblock Masked autoencoders are scalable vision learners.
\newblock In \emph{The Conference on Computer Vision and Pattern Recognition,
  {CVPR}}, 2022.

\bibitem[Krizhevsky et~al.(2009)Krizhevsky, Hinton, and
  et~al.]{krizhevsky2020cifar}
Alex Krizhevsky, Geoffrey Hinton, and et~al.
\newblock Learning multiple layers of features from tiny images.
\newblock 2009.

\bibitem[Le et~al.(2011)Le, Karpenko, Ngiam, and Ng]{le2011ica}
Quoc Le, Alexandre Karpenko, Jiquan Ngiam, and Andrew Ng.
\newblock Ica with reconstruction cost for efficient overcomplete feature
  learning.
\newblock \emph{Advances in Neural Information Processing Systems, {NeurIPS}},
  2011.

\bibitem[Li et~al.(2022)Li, Chen, LeCun, and Sommer]{li2022neural}
Zengyi Li, Yubei Chen, Yann LeCun, and Friedrich~T Sommer.
\newblock Neural manifold clustering and embedding.
\newblock \emph{arXiv preprint arXiv:2201.10000}, 2022.

\bibitem[Mikolov et~al.(2013{\natexlab{a}})Mikolov, Chen, Corrado, and
  Dean]{mikolov2013efficient}
Tom{\'{a}}s Mikolov, Kai Chen, Greg Corrado, and Jeffrey Dean.
\newblock Efficient estimation of word representations in vector space.
\newblock In \emph{The International Conference on Learning Representations,
  {ICLR} Workshop Track Proceedings}, 2013{\natexlab{a}}.

\bibitem[Mikolov et~al.(2013{\natexlab{b}})Mikolov, Sutskever, Chen, Corrado,
  and Dean]{mikolov2013distributed}
Tomas Mikolov, Ilya Sutskever, Kai Chen, Greg~S Corrado, and Jeff Dean.
\newblock Distributed representations of words and phrases and their
  compositionality.
\newblock \emph{Advances in Neural Information Processing Systems, {NeurIPS}},
  2013{\natexlab{b}}.

\bibitem[Pennington et~al.(2014)Pennington, Socher, and
  Manning]{pennington2014glove}
Jeffrey Pennington, Richard Socher, and Christopher~D Manning.
\newblock Glove: Global vectors for word representation.
\newblock In \emph{Proceedings of the Conference on Empirical Methods in
  Natural Language Processing, {EMNLP}}, 2014.

\bibitem[Roweis \& Saul(2000)Roweis and Saul]{roweis2000nonlinear}
Sam~T Roweis and Lawrence~K Saul.
\newblock Nonlinear dimensionality reduction by locally linear embedding.
\newblock \emph{Science}, 2000.

\bibitem[Rumelhart et~al.(1986)Rumelhart, Hinton, and
  Williams]{rumelhart1986learning}
David~E Rumelhart, Geoffrey~E Hinton, and Ronald~J Williams.
\newblock Learning representations by back-propagating errors.
\newblock \emph{Nature}, 1986.

\bibitem[Tenenbaum et~al.(2000)Tenenbaum, De~Silva, and
  Langford]{tenenbaum2000global}
Joshua~B Tenenbaum, Vin De~Silva, and John~C Langford.
\newblock A global geometric framework for nonlinear dimensionality reduction.
\newblock \emph{Science}, 2000.

\bibitem[Thiry et~al.(2021)Thiry, Arbel, Belilovsky, and
  Oyallon]{thiry2021patchinconvkernel}
Louis Thiry, Michael Arbel, Eugene Belilovsky, and Edouard Oyallon.
\newblock The unreasonable effectiveness of patches in deep convolutional
  kernels methods.
\newblock In \emph{The International Conference on Learning Representations,
  {ICLR}}, 2021.

\bibitem[Trockman \& Kolter(2023)Trockman and Kolter]{trockman2022patches}
Asher Trockman and J.~Zico Kolter.
\newblock Patches are all you need?
\newblock \emph{Transactions on Machine Learning Research, {TMLR}}, 2023.

\bibitem[Wiskott \& Sejnowski(2002)Wiskott and Sejnowski]{wiskott2002slow}
Laurenz Wiskott and Terrence~J Sejnowski.
\newblock Slow feature analysis: Unsupervised learning of invariances.
\newblock \emph{Neural Computation}, 2002.

\bibitem[Wu et~al.(2018)Wu, Xiong, Yu, and Lin]{wu2018unsupervised}
Zhirong Wu, Yuanjun Xiong, Stella~X Yu, and Dahua Lin.
\newblock Unsupervised feature learning via non-parametric instance
  discrimination.
\newblock In \emph{Proceedings of the IEEE Conference on Computer Vision and
  Pattern Recognition, {CVPR}}, 2018.

\bibitem[Yeh et~al.(2022)Yeh, Hong, Hsu, Liu, Chen, and
  LeCun]{yeh2021decoupled}
Chun{-}Hsiao Yeh, Cheng{-}Yao Hong, Yen{-}Chi Hsu, Tyng{-}Luh Liu, Yubei Chen,
  and Yann LeCun.
\newblock Decoupled contrastive learning.
\newblock In \emph{The European Conference on Computer Vision, {ECCV}}, 2022.

\bibitem[Zbontar et~al.(2021)Zbontar, Jing, Misra, LeCun, and
  Deny]{zbontar2021barlow}
Jure Zbontar, Li~Jing, Ishan Misra, Yann LeCun, and St{\'e}phane Deny.
\newblock Barlow twins: Self-supervised learning via redundancy reduction.
\newblock In \emph{International Conference on Machine Learning, {ICML}}, 2021.

\end{thebibliography}
\bibliographystyle{tmlr}

\clearpage

\appendix
\section*{Appendix}

\section{Proof of Proposition~\ref{prop:cooccurrence}}
\label{app:proof_1}

\begin{proof}
Since we are dealing with an objective, we can drop constants, which do not depend on the embedding $\vec{z}_1$ and $\vec{z}_2$, when they occur.
\begin{align}
L&=\int p(\vec{x}_1)p(\vec{x}_2) \left[ w\vec{z}_1^T\vec{z}_2 - \frac{p(\vec{x}_1, \vec{x}_2)}{p(\vec{x}_1)p(\vec{x}_2)} \right]^2 d\vec{x}_1 d\vec{x}_2 \\
&= \int p(\vec{x}_1)p(\vec{x}_2) \left[ (w\vec{z}_1^T\vec{z}_2)^2 - 2w\vec{z}_1^T\vec{z}_2\cdot \frac{p(\vec{x}_1, \vec{x}_2)}{p(\vec{x}_1)p(\vec{x}_2)} \right]  d\vec{x}_1 d\vec{x}_2\\
&= \int p(\vec{x}_1)p(\vec{x}_2) (w\vec{z}_1^T\vec{z}_2)^2 d\vec{x}_1 d\vec{x}_2 - 2w\int p(\vec{x}_1,\vec{x}_2) (\vec{z}_1^T\vec{z}_2) d\vec{x}_1 d\vec{x}_2\\
&= \mathbb{E}_{p(\vec{x}_1, \vec{x}_2)}\left[- \vec{z}_1^T\vec{z}_2\right] + \lambda\mathbb{E}_{p(\vec{x}_1)p(\vec{x}_2)} \left(\vec{z}_1^T\vec{z}_2\right)^2
\end{align}
where $\lambda=\frac{w}{2}$. 
\end{proof}
In practice, we can simply choose $w = 2$ so that $\lambda=1$.

\section{The Duality Between Contrastive and Non-Contrastive SSL}
\label{app:duality_short}
The similarity term in different joint-embedding methods is essentially the same, and we focus on the regularization term, particularly with SGD optimizer. For simplicity, we assume that the embedding $\vec{z}$ is L2-normalized and each of the embedding dimension also has zero mean and normalized variance. Given a minibatch with size $N$, the spectral regularization term $\mathbb{E}_{p(\vec{x}_1)p(\vec{x}_2)} \left(\vec{z}_1^T\vec{z}_2\right)^2$ reduces to $\left\|Z^T Z - I_d\right\|_F^2$. By Lemma 3.2 from \citet{le2011ica}, we have: 
\begin{align}
\left\|Z^T Z - I_N\right\|_F^2 = \left\|ZZ^T - I_d\right\|_F^2 = \left\|ZZ^T - \frac{N}{d} I_d\right\|_F^2 + C
\end{align}
where $C$ is a constant. The third equality follows due to that each of the embedding dimension is normalized. $\left\|ZZ^T - \frac{1}{d} I_N\right\|_F^2$ is the mini-batch version of the covariance regularization term $\mathbb{E}_{p(\vec{x})}\left[ \vec{z} \vec{z}^T\right]  =  \frac{N}{d_{emb}}\cdot I$. For a thorough discussion on the duality between contrastive learning and non-contrastive learning, we refer the curious readers to \citep{Garrido2023duality}.

\section{Implementation Details}
\label{appendix: Details}
\subsection{CIFAR-10 and CIFAR-100}
For all experiments and methods we pretrain a ResNet-34 for 600 epochs. We use a batch size of 1024, LARS optimizer, and a weight decay of $1e-04$. The learning rate is set to 0.3, and follows a cosine decay schedule, with 10 epochs of warmup and a final value of 0. In the TCR loss, $\lambda$ is set to 30.0, and $\epsilon$ is set to 0.2. In the SimCLR loss, we set the temperature to 0.1, In the VICReg loss, we set the coefficients for variance/invariance/covariance to 25, 25, 1 respectively. For all methods, the projector network consists of 2 linear layers with respectively 4096 hidden units and 128 output units for the CIFAR-10 experiments and 512 output units for the CIFAR-100 experiments. All the layers are separated with a ReLU and a BatchNorm layers. For all the methods, the data augmentations used are identical to those described in BYOL~\cite{grill2020BYOL}.

\subsection{ImageNet-100}
For all the experiments and methods we pretrain a ResNet-50 for 400 epochs. We use a batch size of 1024, the LARS optimizer, and a weight decay of $1e-04$. The learning rate is set to 0.1, and follows a cosine decay schedule, with 10 epochs of warmup and a final value of 0. In the TCR loss, $\lambda$ is set to 1920.0, and $\epsilon$ is set to 0.2. In the SimCLR loss, we set the temperature to 0.1, In the VICReg loss, we set the coefficients for variance/invariance/covariance to 25, 25, 1 respectively. For all methods, the projector network consists of 3 linear layers with each and 8192 units, separated by a ReLU and a BatchNorm layers. For all the methods, the data augmentations used are identical to those described in BYOL~\cite{grill2020BYOL}.

\subsection{ImageNet}
For all the experiments we pretrain a ResNet-50 for 800 epochs with VICReg~\cite{bardes2021vicreg}. We use a batch size of 1024, the LARS optimizer, and a weight decay of $1e-06$. The learning rate is set to 0.1, and follows a cosine decay schedule, with 10 epochs of warmup and a final value of 0. We set the coefficients for variance/invariance/covariance to 25, 25, 1 respectively. The projector network consists of 3 linear layers with each and 8192 units, separated by a ReLU and a BatchNorm layers. The data augmentations used are identical to those described in BYOL~\cite{grill2020BYOL}.

\subsection{Implementation Detail for Local Aggregation}
For all the experiments, we downloaded the checkpoints of SOTA SSL model pretrained on CIFAR10 dataset from \href{https://github.com/vturrisi/solo-learn}{solo-learn}. Each method is pretrained for 1000 epochs and the hyperparameters used for each method is described in \href{https://github.com/vturrisi/solo-learn}{solo-learn}. The backbone model used in all these checkpoint is ResNet-18, which output a dimension $512 \times 5 \times 5$ tensor for each image. We apply spatial average pooling (stride = 2, window size = 3) to this tensor and flatten the result to obtain a feature vector of dimension $2048$.

\section{ImageNet Visualization}
\begin{figure}[h]
\begin{center}
\centerline{\includegraphics[width=0.75\columnwidth]{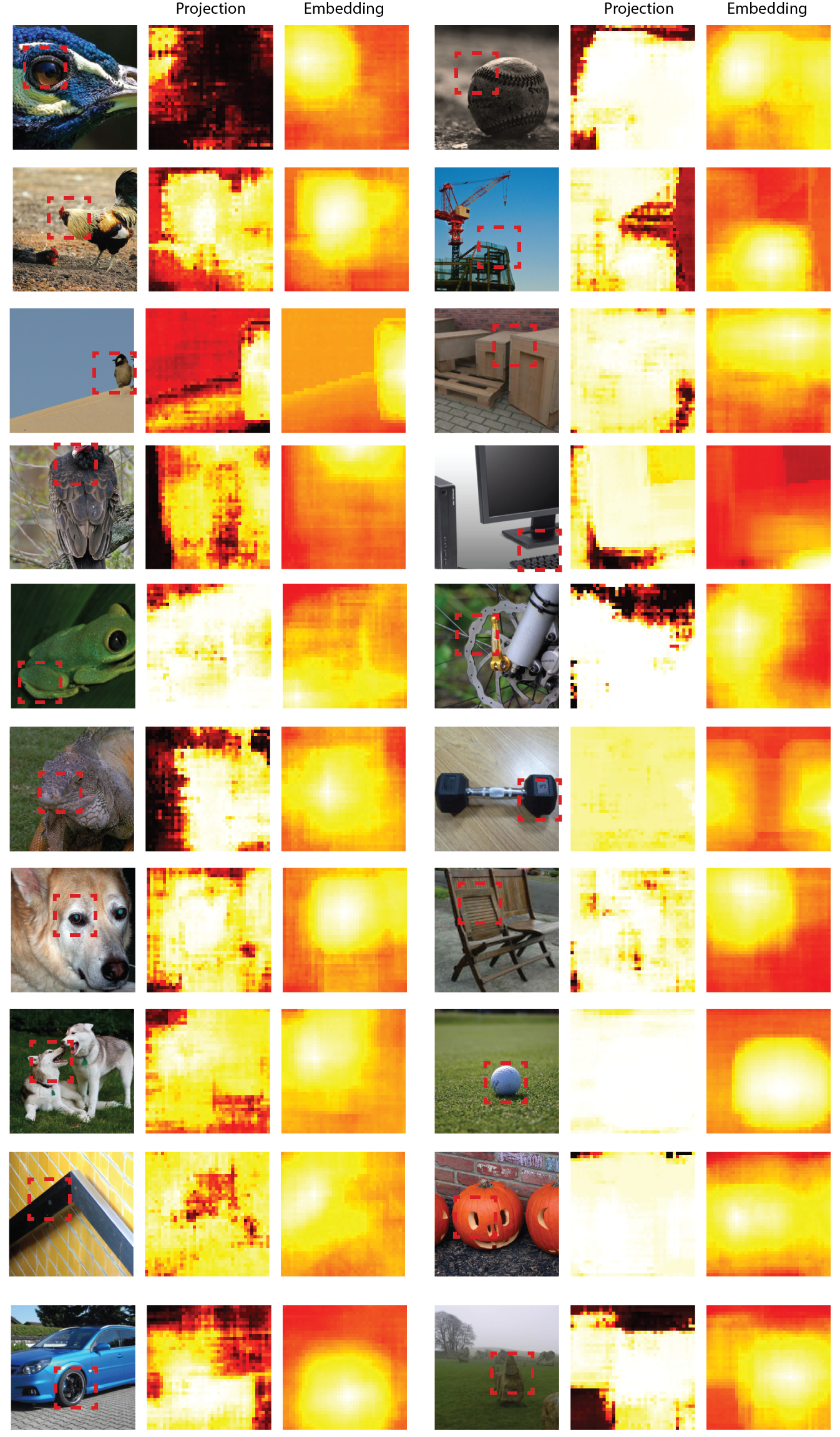}}
\end{center}
\vspace{-0.3in} 
\caption{\textbf{More visualization of cosine similarity heatmaps in the projection space and the embedding space.} Here the query patch is marked by the red-dash boxes and its size is $71\times 71$ and the instance image size is $224\times 224.$}
\label{sup:imagenet_heatmaps}
\end{figure}

In this section, we provide further visualization of the multi-scale pretrained VICReg model, and the results are shown in Fig~\ref{sup:imagenet_heatmaps}. Here we use image patches of scale $0.1$ to calculate the cosine similarity heatmaps, the query patch is marked by the red-dash boxes. The embedding space contains more localized information, whereas the projection space is relatively more invariant, especially when the patch has enough information to determine the category.

\section{CIFAR10 kNN Visualization}
\label{app:visualization}
This section continues additional visualization of the model pretrained with $14\times 14$ patches on CIFAR-10. In this visualization, we use cosine similarity to find the closest neighbors for the query patches, marked in the red-dash boxes. Again, green boxes indicate that the patches are from other images of the same category; red boxes indicate that the patches are from other images of a different category. Patches that do not have a colored box are from the same image. In the following, we discuss several aspects of the problem.

\begin{figure}[h]
\begin{center}
\centerline{\includegraphics[width=1.0\columnwidth]{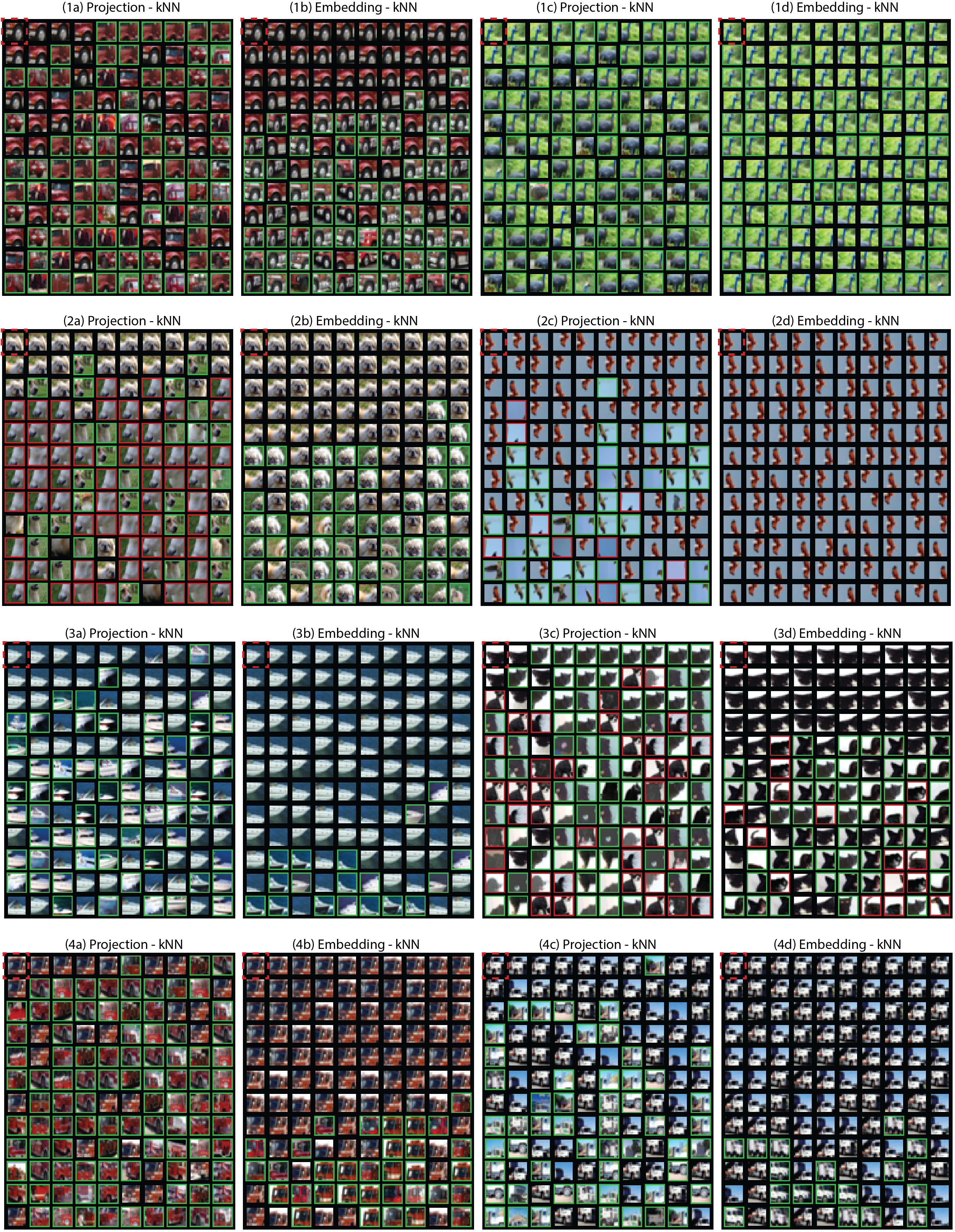}}
\end{center}
\vspace{-0.3in} 
\caption{\textbf{Additional comparison between the projection space and the embedding space.}}
\label{sup:proj_emb_knn}
\end{figure}

{\bf Additional Projection and Embedding Spaces Comparison.} As we can see in Figure~\ref{sup:proj_emb_knn}, the embedding space has a lesser degree of information collapse. The projection space tends to collapse different ``parts'' of a class to similar vectors, whereas the embedding space preserves more information about the locality or details in a patch. This is manifested by higher visual similarity between neighboring patches.

\begin{figure}[h]
\begin{center}
\centerline{\includegraphics[width=1.0\columnwidth]{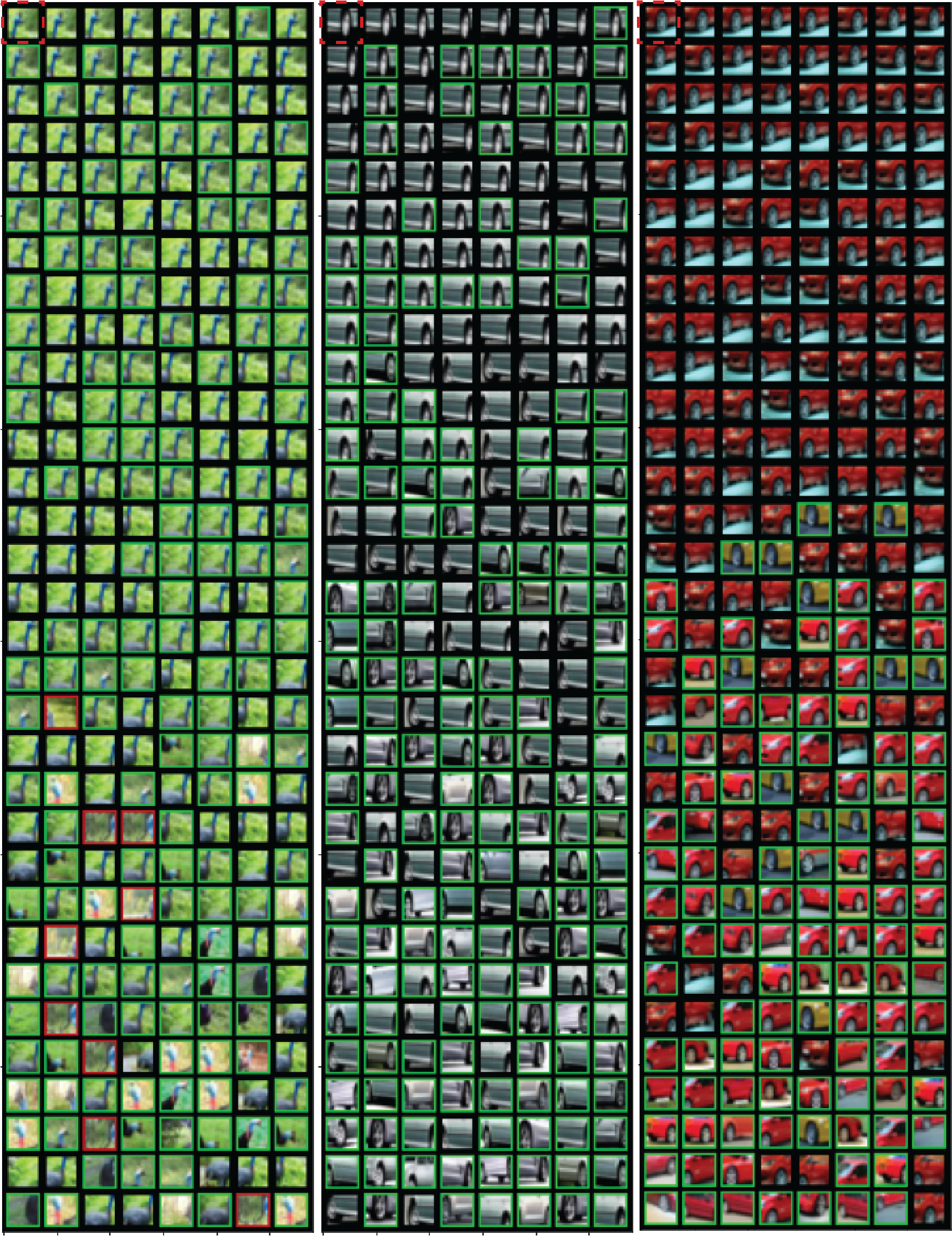}}
\end{center}
\vspace{-0.3in} 
\caption{\textbf{kNN in the embedding Space with 255 neighbors.}}
\label{sup:knn_patch_1}
\end{figure}

\begin{figure}[h]
\begin{center}
\centerline{\includegraphics[width=1.0\columnwidth]{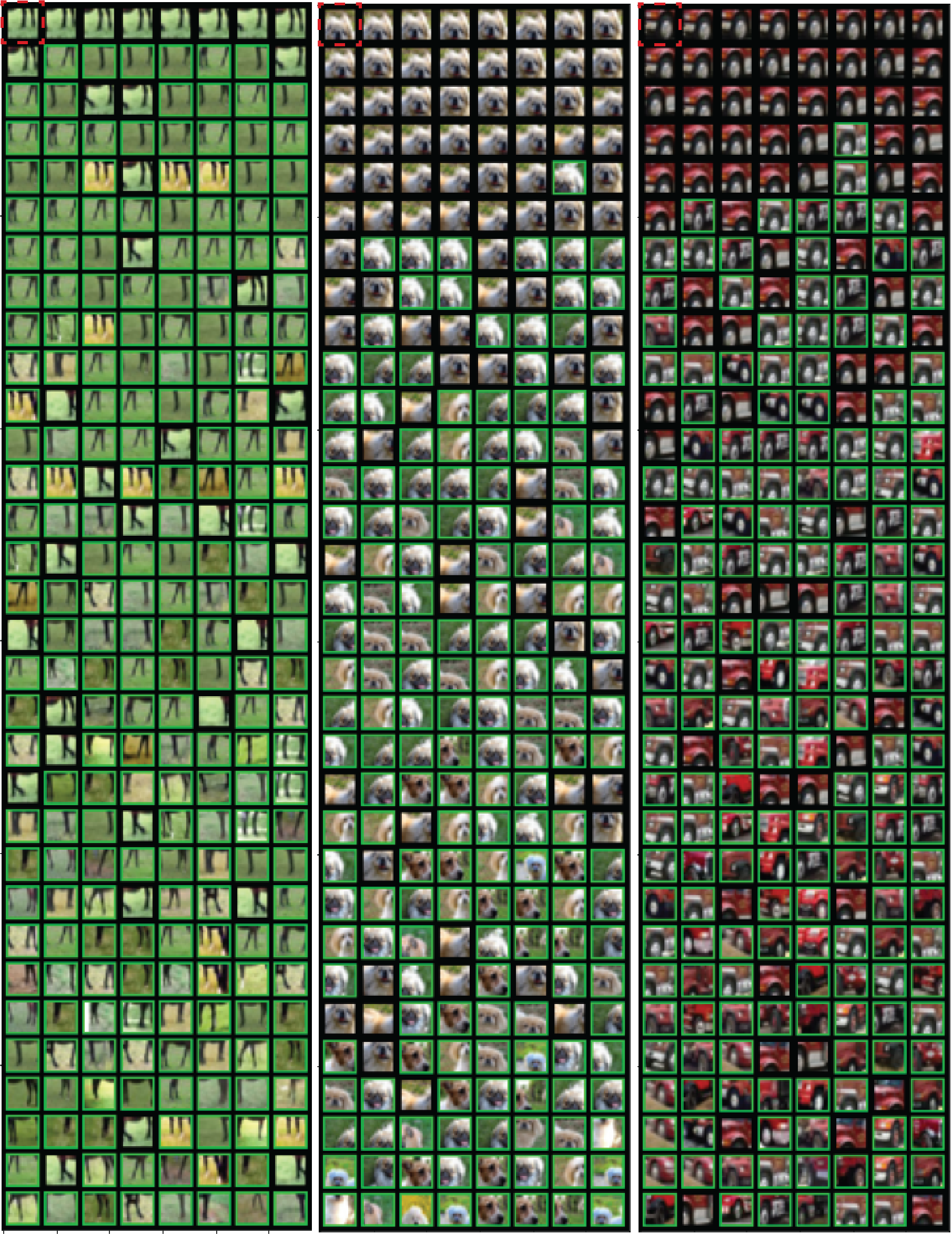}}
\end{center}
\vspace{-0.3in} 
\caption{\textbf{kNN in the embedding Space with 255 neighbors.}}
\label{sup:knn_patch_2}
\end{figure}

{\bf Embedding Space with 256 kNN.} In the previous CIFAR visualization, we only show kNN with $119$ neighbors. In Figure~\ref{sup:knn_patch_1} and Figure~\ref{sup:knn_patch_2}, we provide kNN with $255$ neighbors, and the same conclusions hold.

\begin{figure}[h]
\begin{center}
\centerline{\includegraphics[width=1.0\columnwidth]{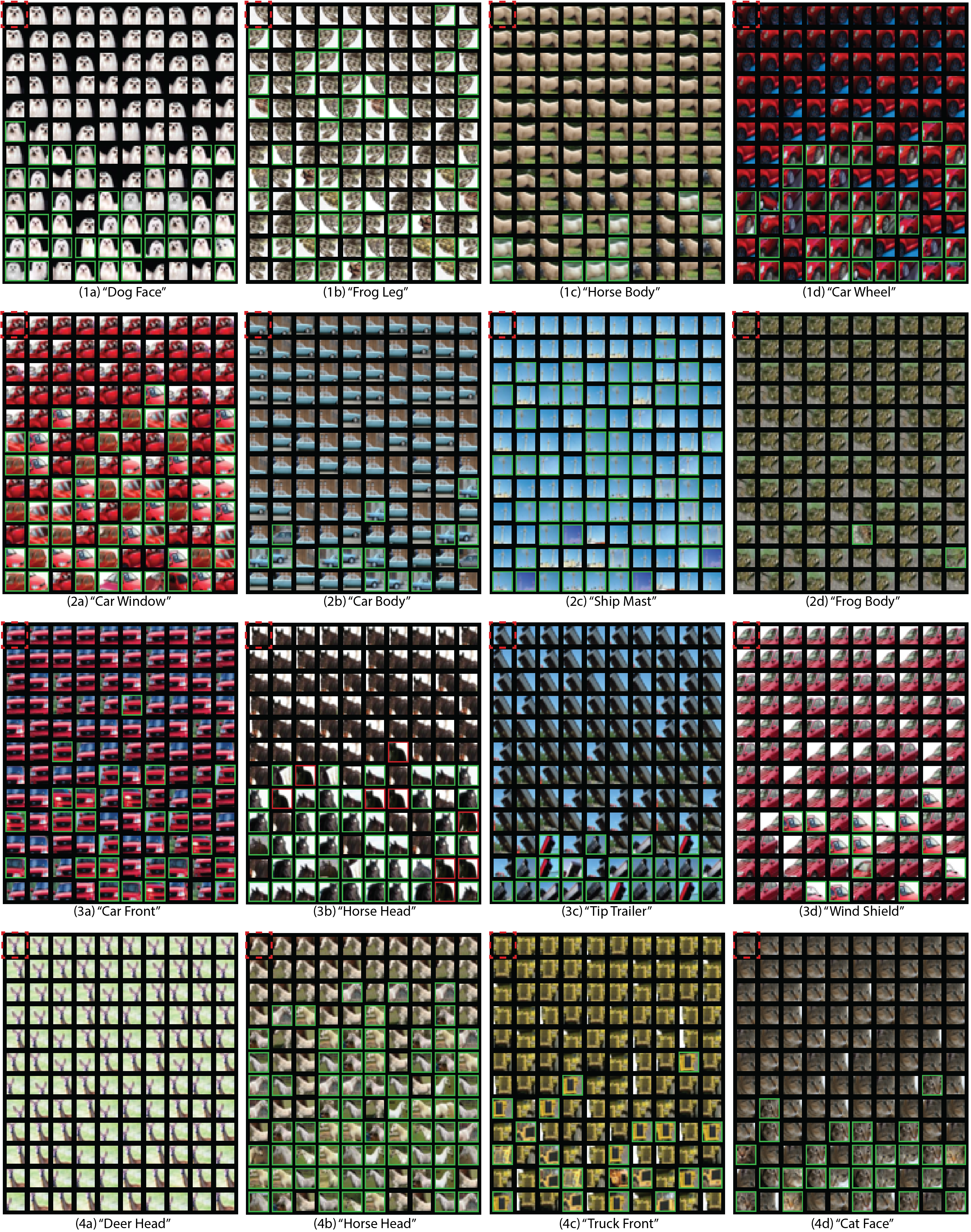}}
\end{center}
\vspace{-0.3in} 
\caption{\textbf{Different ``parts'' in the embedding space.}}
\label{sup:emb_parts}
\end{figure}

{\bf Different ``Parts'' in the Embedding Space.} In Figure~\ref{sup:emb_parts}, we provide more typical patches of ``parts'' and show their embedding neighbors. While many parts are shared by different instances, we also find several less ideal cases, e.g., Figure~\ref{sup:emb_parts}(4a)(2d), where the closest neighbors are nearly all from the same image.

\begin{figure}[h]
\begin{center}
\centerline{\includegraphics[width=0.96\columnwidth]{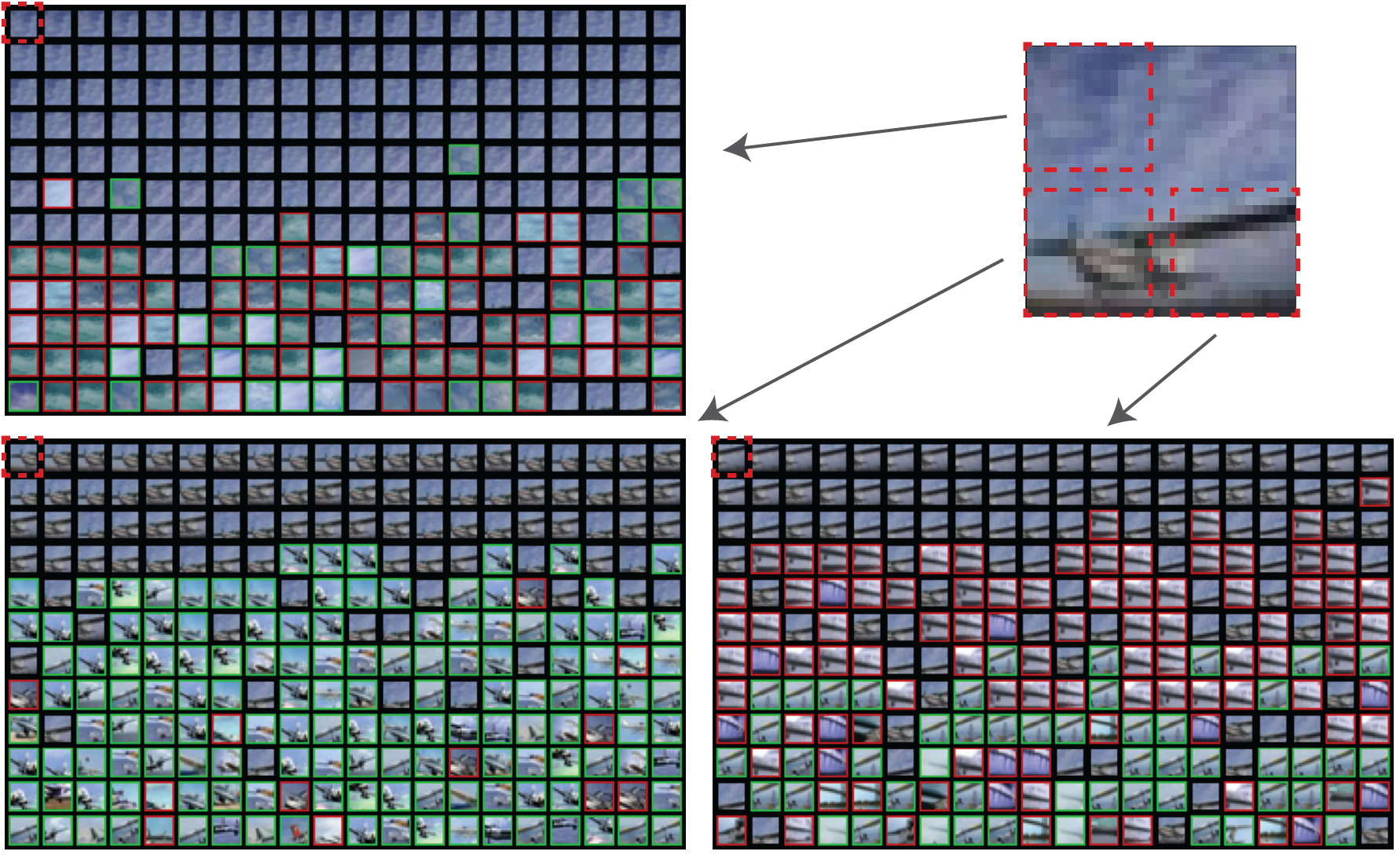}}
\end{center}
\vspace{-0.3in} 
\caption{\textbf{The compositional structure of an airplane.} The ``sky'' part is shared by ships, birds, etc. The ``wing'' resembles the silhouettes of ships and is also shared by flying birds. The airscrew part is primarily shared by the other airplanes.}
\label{sup:instance_parts_1}
\end{figure}

\begin{figure}[h]
\begin{center}
\centerline{\includegraphics[width=0.96\columnwidth]{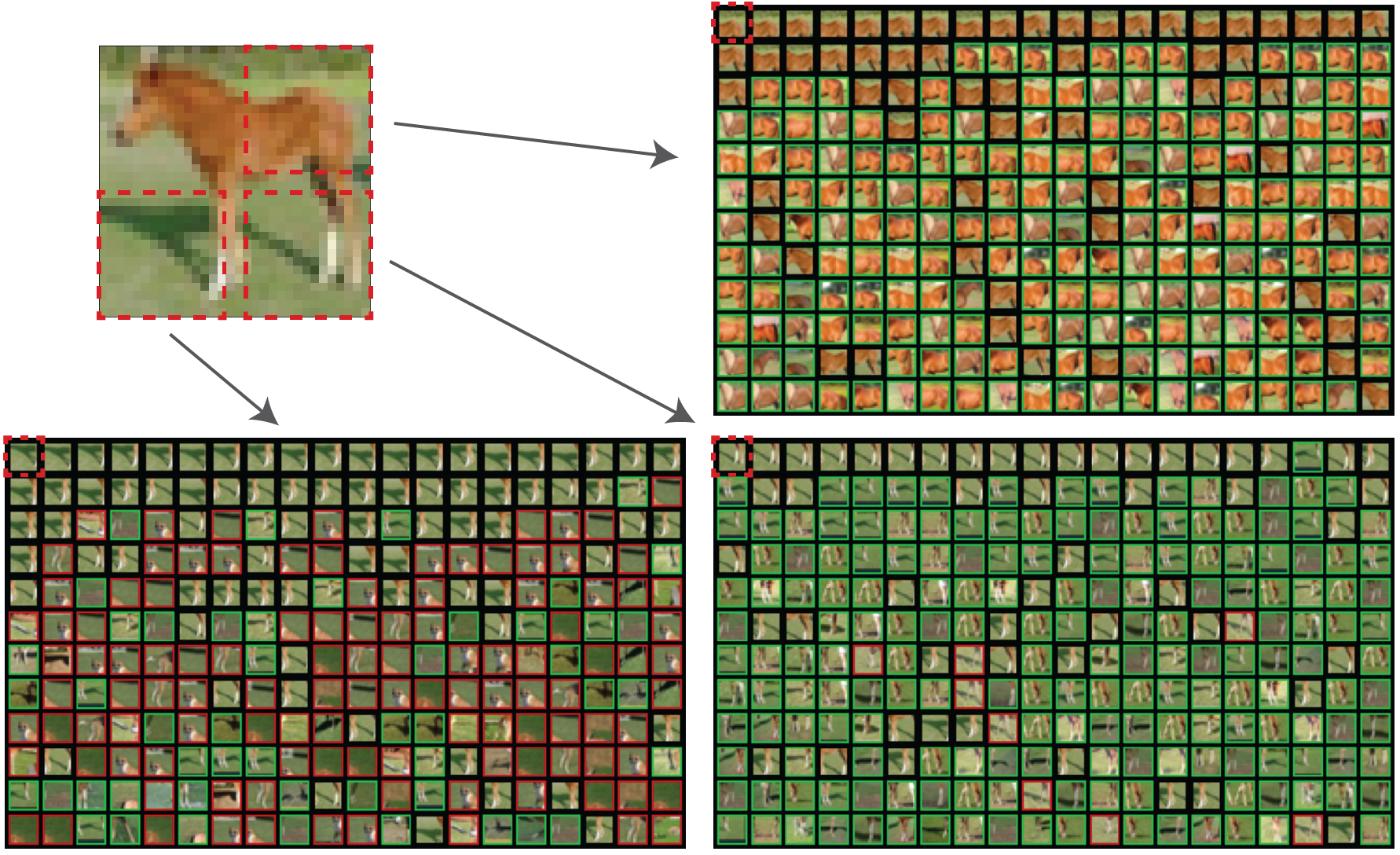}}
\end{center}
\vspace{-0.3in} 
\caption{\textbf{The compositional structure of a horse.} The bottom left corner contains ``shadow'', and the similar shadows are shared by deers and dogs. The bottom right part contains ``legs'', which are also shared by deers and dogs. However, from the back to the thigh is shared by primarily other horses.}
\label{sup:instance_parts_2}
\end{figure}

As we discussed earlier, the objective is modeling patches' co-occurrence statistics. It is relatively uninformative if the same patch is not ``shared'' by different instances. While the same patch might not be ``shared'', the color augmentation and deep image prior embedded in the network design may create approximate sharing. In Figure~\ref{sup:instance_parts_1} and Figure~\ref{sup:instance_parts_2}, we provide two examples of the compositional structure of instances.

\end{document}